\PassOptionsToPackage{prologue,dvipsnames}{xcolor}
\documentclass[letterpaper]{article} % DO NOT CHANGE THIS
\usepackage{aaai25}  % DO NOT CHANGE THIS
\usepackage{times}  % DO NOT CHANGE THIS
\usepackage{helvet}  % DO NOT CHANGE THIS
\usepackage{courier}  % DO NOT CHANGE THIS
\usepackage[hyphens]{url}  % DO NOT CHANGE THIS
\usepackage{graphicx} % DO NOT CHANGE THIS
\usepackage{natbib}  % DO NOT CHANGE THIS AND DO NOT ADD ANY OPTIONS TO IT
\usepackage{caption} % DO NOT CHANGE THIS AND DO NOT ADD ANY OPTIONS TO IT
\usepackage{algorithm}
\usepackage{algorithmic}

\usepackage{newfloat}
\usepackage{listings}
\DeclareCaptionStyle{ruled}{labelfont=normalfont,labelsep=colon,strut=off} % DO NOT CHANGE THIS
\floatstyle{ruled}
\newfloat{listing}{tb}{lst}{}
\floatname{listing}{Listing}
\usepackage{booktabs}
\usepackage{multicol}
\usepackage{multirow}
\usepackage{tcolorbox}
\usepackage{colortbl}
\usepackage{util}
\usepackage[dvipsnames]{xcolor}

\lstdefinestyle{python}{
    language=Python,
    basicstyle=\fontsize{7}{9.5}\ttfamily,
    keywordstyle=\color{blue},
    commentstyle=\color{gray},
    stringstyle=\color{black},
    showstringspaces=false,
    breaklines=true,
    breakindent=0pt,
    breakatwhitespace=false,
    escapeinside={(*@}{@*)}
}

\lstdefinestyle{plain}{
    basicstyle=\fontsize{7}{9.5}\ttfamily,
    keywordstyle=\color{blue},
    commentstyle=\color{gray},
    stringstyle=\color{green},
    showstringspaces=false,
    breaklines=true,
    breakatwhitespace=false,
    breakindent=0pt,
    escapeinside={(*@}{@*)}
}

\title{Template-Driven LLM-Paraphrased Framework for Tabular \\ Math Word Problem Generation}
\author{
    Xiaoqiang Kang\textsuperscript{\rm 1,2,}\equalcontrib,
    Zimu Wang\textsuperscript{\rm 1,2,}\equalcontrib,
    Xiaobo Jin\textsuperscript{\rm 1},
    Wei Wang\textsuperscript{\rm 1},
    Kaizhu Huang\textsuperscript{\rm 3},
    Qiufeng Wang\textsuperscript{\rm 1,}\thanks{Corresponding author.}
}
\affiliations{
    \textsuperscript{\rm 1}Xi'an Jiaotong-Liverpool University \quad
    \textsuperscript{\rm 2}University of Liverpool \quad
    \textsuperscript{\rm 3}Duke Kunshan University \\
    \{xiaoqiang.kang23, zimu.wang19\}@student.xjtlu.edu.cn \\
    \{xiaobo.jin, wei.wang03, qiufeng.wang\}@xjtlu.edu.cn \\
    kaizhu.huang@dukekunshan.edu.cn
}

\usepackage{bibentry}
\begin{document}

\maketitle

\begin{abstract}
Solving tabular math word problems (TMWPs) has become a critical role in evaluating the mathematical reasoning ability of large language models (LLMs), where large-scale TMWP samples are commonly required for LLM fine-tuning. Since the collection of high-quality TMWP datasets is costly and time-consuming, recent research has concentrated on automatic TMWP generation. However, current generated samples usually suffer from issues of either correctness or diversity. In this paper, we propose a \textbf{Te}mplate-driven \textbf{LL}M-paraphrased (TeLL) framework for generating high-quality TMWP samples with diverse backgrounds and accurate tables, questions, answers, and solutions. To this end, we first extract templates from existing real samples to generate initial problems, ensuring correctness. Then, we adopt an LLM to extend templates and paraphrase problems, obtaining diverse TMWP samples. Furthermore, we find the reasoning annotation is important for solving TMWPs. Therefore, we propose to enrich each solution with illustrative reasoning steps. Through the proposed framework, we construct a high-quality dataset TabMWP-TeLL by adhering to the question types in the TabMWP dataset, and we conduct extensive experiments on a variety of LLMs to demonstrate the effectiveness of TabMWP-TeLL in improving TMWP solving performance. The code and data of this paper are available at: https://github.com/Jason8Kang/TELL.
\end{abstract}


\section{Introduction}

The rise of large language models (LLMs) has achieved unprecedented success in a variety of reasoning tasks \cite{peng2023doesincontextlearningfall,li2024meqa,wang-etal-2024-document-level}. However, solving math word problems (MWPs) is still challenging, which is tasked as answering math questions based on heterogeneous tabular and textual data with mathematical reasoning ability \cite{lu2023dynamic,zheng-etal-2023-chain}. For various complex MWPs, training models usually require a large amount of data. Nevertheless, the collection and annotation of MWPs are usually costly and time-consuming, resulting in the scarcity of public tabular MWP datasets.

To mitigate the data issue, numerous studies have explored the ability to automatically generate MWP samples, mainly including template-based methods \cite{williams2011generating,polozov2015personalized}, rewriting-based methods \cite{koncel-kedziorski-etal-2016-theme}, neural network-based methods \cite{liyanage-ranathunga-2020-multi,liu-etal-2021-mathematical} and LLM-based methods \cite{luo2023wizardmathempoweringmathematicalreasoning,tang2024mathscale}. Despite the progress, existing MWP generation methods still face three major challenges for tabular data:

\begin{figure}[t!]
    \centering
    \includegraphics[width=\linewidth]{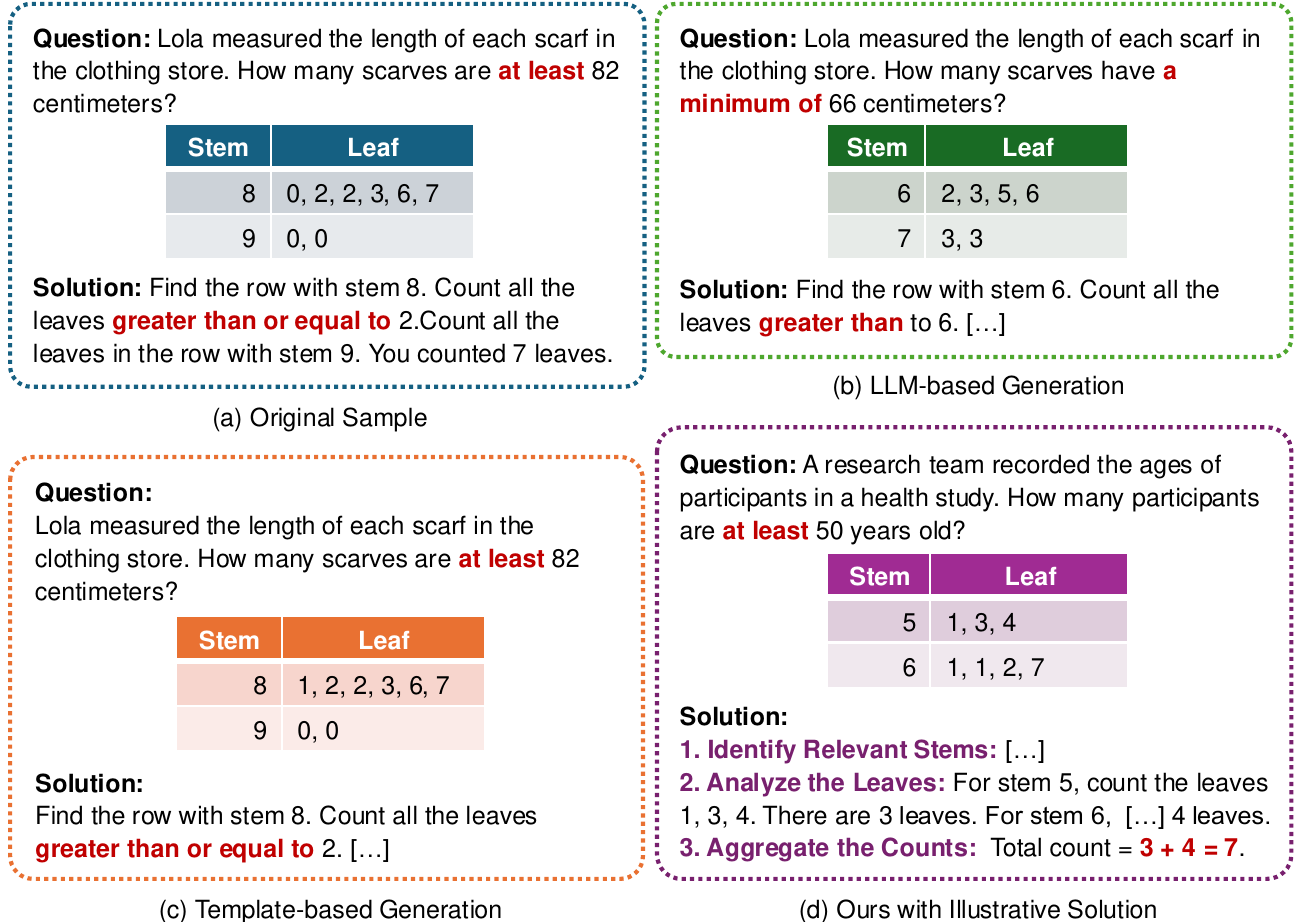}
    \caption{Illustration of an original TMWP sample and the generated samples with LLMs, templates, and ours with an illustrative solution. (Notes: ``Stem'' means the first digit, and ``Leaf'' means the last digit in stem-leaf plots).}
    \label{fig:background}
\end{figure}

(1) \textbf{Lack of correctness}. Generation-based methods, like rewriting-based and LLM-based methods, misunderstand the meaning of the questions due to the hallucination problem \cite{zhang2023sirenssongaiocean} and thus get wrong answers. As shown in Figure \ref{fig:background}(b), the LLM-generated solution calculates leaves greater than $66$ but ignores the case equal to $66$, which is inconsistent with the requirements of the question. 
(2) \textbf{Lack of diversity in problems}. Because all problems are generated from abstract templates \cite{williams2011generating}, template-based methods have limited diversity. As shown in Figure \ref{fig:background}(c), the generated problem simply replaces some numbers that do not affect the answer while keeping the overall content unchanged. 
(3) \textbf{Lack of illustrative steps in solutions}. As shown in Figures \ref{fig:background}(a) and \ref{fig:background}(d), our data annotations have more clearly described solution steps than other data annotations. The model induces multi-step reasoning behaviors through clear intermediate reasoning steps such as Chain-of-Thought (CoT) \cite{NEURIPS2022_9d560961}.

\begin{figure*}[t!]
    \centering
    \includegraphics[width=\linewidth]{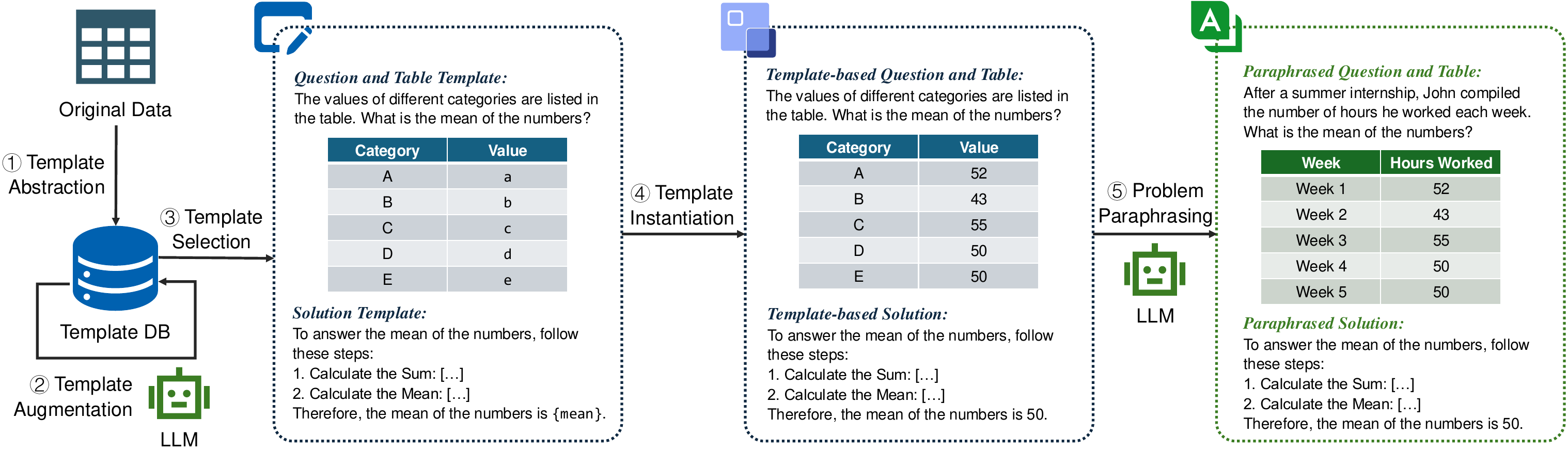}
    \caption{Overall framework of the proposed TeLL method to generate TMWPs with correctness and diversity, consisting of five steps: 1) template abstraction, 2) template augmentation, 3) template selection, 4) template instantiation, and 5) problem paraphrasing.}
    \label{fig:framework}
\end{figure*}

To overcome the aforementioned challenges, we propose a \textbf{Te}mplate-driven \textbf{LL}M Paraphrased (TeLL) framework for generating Tabular MWPs (TMWPs) using both templates and LLMs. Different from the previous template-based generation, which rewrites questions with minor modifications based on a pre-defined template as shown in Figure \ref{fig:background}(c), our templates are extracted from an existing TMWP dataset, each of which is abstract and summarizes mathematical logic. To generate flexible templates, we utilize an LLM to extend those extracted templates with a broader range of question types, maintaining mathematical logic to ensure correctness. Although these extended templates can generate various TMWPs, the background and linguistic description of those generated problems are monotonous. To pursue high-quality, realistic generated samples, we leverage the powerful language ability of LLMs to paraphrase problems with various contexts, obtaining diverse problems. Since our LLM-based paraphrasing does not change the mathematical logic, the correctness can be ensured. The overall framework is shown in Figure \ref{fig:framework}, and the details can be found in the section of Methodology. In summary, our generation method combines the advantages of both templates and LLM, ensuring the correctness and diversity of the generated samples.

Based on the proposed framework, we construct a high-quality dataset named TabMWP-TeLL based on the question types in the TabMWP dataset \cite{lu2023dynamic}. We find that the step-by-step reasoning annotations are significant for using LLMs; therefore, we propose refining the original solutions with more illustrative steps, as shown in Figure \ref{fig:background}(d). In our generated dataset, we utilize an LLM, Yi \cite{ai2024yi}, to paraphrase the template-based problems. In experiments, we fine-tune three LLMs, including Mistral \cite{jiang2023mistral7b}, Qwen 2 \cite{yang2024qwen2technicalreport}, and Llama 3 \cite{dubey2024llama3herdmodels}. Experimental results show that TabMWP-TeLL is effective in improving TMWP solving, outperforming the baselines by a large margin, and is particularly effective in improving performance on challenging problems while maintaining the performance on simple ones.

Our contributions can be summarized as follows:

\begin{itemize}
    \item We propose TeLL, a template-driven, LLM-paraphrased framework, to generate high-quality TMWPs. To the best of our knowledge, we are the first to leverage templates and LLMs on TMWP generation, ensuring both correctness and diversity. 
    \item We propose to enrich TMWP solutions with more illustrative annotations, eliciting the multi-step reasoning ability of LLMs.
    \item We construct a high-quality TMWP dataset, TabMWP-TeLL, which is an extension of the TabMWP dataset. The results of human verification illustrate certain correctness and diversity of the TMWP generation strategy.
    \item Extensive experimental results demonstrate the effectiveness of TabMWP-TeLL in improving TMWP solving, outperforming the baselines by a significant margin.
\end{itemize}

\section{Related Work}
\label{sec:related}

\paragraph{Math Word Problems.} Recent research has primarily focused on addressing MWPs using generative models, such as sequence-based \cite{wang-etal-2017-deep} and tree-based \cite{ijcai2019p736,zhang-etal-2020-graph-tree} models. With the rapid advancements of LLMs and the development of few-shot \cite{NEURIPS2020_1457c0d6} and CoT \cite{NEURIPS2022_9d560961} prompting, there has been a growing trend toward leveraging prompt engineering \cite{chen2023program,fu2023complexitybased,zhou2023leasttomost,wang-etal-2023-plan} and fine-tuning \cite{liu2023improvinglargelanguagemodel,liang-etal-2024-mint} strategies with notable performance. Furthermore, LLMs with math reasoning abilities have also been incorporated in the field of intelligent education \cite{macina-etal-2023-mathdial,wang-demszky-2023-chatgpt}. In this context, generating high-quality MWPs with both correctness and diversity is worthwhile.

\paragraph{Tabular Math Word Problems.} Recent years have witnessed extensive research into solving TMWPs. TabMWP \cite{lu2023dynamic} is the first TMWP reasoning dataset that contains $38,431$ grade-level problems with tabular context. PromptPG has been subsequently proposed, which utilizes policy gradient to select in-context examples for the test examples. Considering that LLMs often make errors in mathematical calculations, subsequent research has primarily focused on using external tools, such as calculators, to improve calculation accuracy.
TaCo \cite{zheng-etal-2023-chain} coordinates two Tabular LMs (TaLMs), which are responsible for CoT generation and answer inference, integrated with an external calculator.
Chameleon \cite{NEURIPS2023_871ed095} composes various tools to accomplish complex reasoning tasks, such as LLMs, Python functions, and row and column look-up.
CREATOR \cite{qian-etal-2023-creator} and CRAFT \cite{yuan2024craft} create new tools for specific problems rather than calling the existing ones.

\paragraph{Math Word Problem Generation.} Existing research in MWP generation can be broadly classified into four categories, including template-based, rewriting-based, neural network-based, and LLM-based methods. Template-based methods start by abstracting the existing MWPs into a template or a skeleton and then generating new problems from the abstract templates \cite{williams2011generating,polozov2015personalized}.
Rewriting-based methods edit existing MWPs, altering the background of the problems while preserving their contents and logic \cite{koncel-kedziorski-etal-2016-theme,Moon-Gilbe2019xm}.
Neural network-based methods generate MWPs from topics and equations in an end-to-end manner \cite{liyanage-ranathunga-2020-multi,liu-etal-2021-mathematical,zhou-etal-2023-learning-analogy}.
Inspired by the development of LLMs and their notable performance in various downstream applications, recent attempts have been focused on exploiting LLMs to generate MWPs, such as based on concepts \cite{tang2024mathscale} and key points \cite{huang2024keypointdrivendatasynthesisenhancement}.
However, the aforementioned methods usually suffer from issues of either correctness or diversity. Different from the previous approaches, we propose a template-driven, LLM-paraphrased framework to generate high-quality TMWP samples with diverse descriptions and correct answers.

\section{Methodology}
\label{sec:method}

\subsection{Problem Definition}

We define Tabular Math Word Problems (TMWPs) as follows: given a table $t$ containing multiple rows and columns and a question $q$ about the table $t$, where the table could be a visual image, natural language text, or a structured database, our task is to generate a correct answer $a$ that matches the ground truth of the question, derived by solution steps $s$. In our work, we focus on solving TMWPs using LLMs.

Note that the problems we are concerned with, regardless of the questions or the tables, are described in natural language texts; therefore, they usually follow certain rules. Further, for a given class of \textbf{problems}, we can abstract the template $P(\boldsymbol{x}) = (Q(\boldsymbol{x}), T(\boldsymbol{x}), A(\boldsymbol{x}), S(\boldsymbol{x}))$, containing a \textbf{question} template, a \textbf{table} template, an \textbf{answer} template, and a \textbf{solution} template, with respect to the problems, each of which contains placeholders that can be filled in the TMWP generation process. $\boldsymbol{x}$ contains all the numbers and their corresponding categories in the table. Intuitively, for a given template, we can directly replace the placeholders to obtain a new template-based question $q^*$, table $t^*$, answer $a^*$, and solution $s^*$ via the following functions:
\begin{equation}
    q^* = Q(\boldsymbol{x}),\ t^* = T(\boldsymbol{x}),\ a^* = A(\boldsymbol{x}),\ s^* = S(\boldsymbol{x}).
    \label{eq:eq1}
\end{equation}
At the same time, to achieve diverse problem generation, we introduce an LLM to paraphrase the template-based problems to adapt to real-world scenarios. For $q^*$, $t^*$, $a^*$, and $s^*$ mentioned above, we can obtain the corresponding paraphrased question $q$, table $t$, answer $a$, and solution $s$ by the LLM as follows:
\begin{equation}
    (q,t,a,s) = \text{LLM}(q^*, t^*, a^*, s^*).
    \label{eq:eq2}
\end{equation}

\subsection{TeLL for TMWP Generation}

\paragraph{Overview.} Figure \ref{fig:framework} shows how the TeLL framework generates TMWPs while achieving the correctness and diversity of the generated problems. It consists of five stages: 1) \textbf{Template Abstraction}: abstracting the templates of mainstream TMWPs from existing datasets to build a template database; 2) \textbf{Template Augmentation}: using an LLM to expand the database to cover broader question types; 3) \textbf{Template Selection}: randomly selecting a template from the database for instantiation; 4) \textbf{Template Instantiation} (Equation (\ref{eq:eq1})): instantiating the selected template into a problem by assigning random numbers and categories with predefined constraints; 5) \textbf{Problem Paraphrasing} (Equation (\ref{eq:eq2})): with the support of an LLM, rewriting the problem into a problem with different contexts that conforms to human cognition. In the following, we will describe the details of each step.

\paragraph{Template Abstraction.} We first abstract the problems and build a template database of mainstream TMWPs under the guidance of existing datasets. Each template contains a question template $Q(\boldsymbol{x})$, a table template $T(\boldsymbol{x})$, an answer template $A(\boldsymbol{x})$, and a solution template $S(\boldsymbol{x})$. Each of the abstracted templates contains some placeholders with predefined arithmetic operations, which are then filled with specific numbers and categories during the instantiation process, thereby updating the question, table, answer, and solution accordingly. Before abstraction, we first generate the illustrative step-by-step solution, denoted by $\hat{s}$, that corresponds to the original solution $s_0$ within the dataset:
\begin{equation}
\hat{s} = \tm{LLM}(q,t,s_0).
\end{equation}
% \tc{red}{
Afterwards, we select a list of representative question types from the dataset and create a seed template database.

\begin{figure}[t!]
    \begin{tcolorbox}[
    %title=Prompt for Template Generalization, 
    left=2mm,right=1mm,top=0mm, bottom=0mm,colback=white,colframe=NavyBlue]
    \begin{lstlisting}[style=plain]
You are given a math word problem with tabular contents, and your task is to develop a versatile exercise template that can generate a wide array of exercises with a table. Please consider the following guidelines for this assignment:

(1) You can use pandas, numpy, random, etc., or
other packages if necessary.
(2) You should construct a general dataframe and transform it into a human-readable table format by Python.
(3) Generate the functions that are used to create the question, answer, and step-by-step solution based on the created table.

This is an demonstration for <Task 1>:
<Demonstration for Task 1>
Please generate the exercise template for <Task 2>.
    \end{lstlisting}
    \end{tcolorbox}
    \caption{Prompt for template augmentation.}
    \label{fig:generalization}
\end{figure}

\paragraph{Template Augmentation.} Due to the time-consuming nature of generating specific templates for each TMWP type, we propose template augmentation to generalize templates for a certain class of question types (such as questions about average calculation) to other categories with similar characteristics (such as median, mode, and range calculation questions). Specifically, we construct an LLM prompt, as shown in Figure \ref{fig:generalization}, to drive the model to infer templates related to new categories (including questions, tables, answers, and solutions). This method leverages the inherent understanding and generalization capabilities of LLM to increase the diversity of generated TMWPs while improving the efficiency of template creation.

\paragraph{Template Selection and Instantiation.} After the template database is built, we randomly select a template from the database for instantiation. For each generation, we first generate random numbers and their corresponding categories to maintain diversity, where the values of the numbers meet specific constraints for different types of questions, such as integers in a certain interval, non-negative numbers, etc. Then, we use the generated numbers and categories to fill in the question and table, calculate the answer, and finally, fill them into the illustrative solution.

\paragraph{Problem Paraphrasing.}

\begin{figure}[t!]
    \begin{tcolorbox}[
    %title=Prompt for Natural Question Paraphrasing, 
    left=2mm,right=1mm,top=0mm, bottom=0mm,colback=white,colframe=OliveGreen]
    \begin{lstlisting}[style=plain]
You need to rewrite the given math word problem to increase data diversity and semantic richness, whose
questions and solutions are generated according to a
uniform template. You should keep the original problem, data, and solution logic unchanged. Specific requirements are as follows:

(1) Question (`question`): You need to add a background to the question, set the problem in a specific scenario before introducing the question, and rewrite the question without changing its original meaning.

(2) Solution (`solution`): Since a background has been introduced to the question, the solution process should also be correspondingly rewritten, but the idea and logic should remain the same. You can appropriately modify any unreasonable parts in the reasoning process based on the actual situation.

(3) Table Content (`table_for_pd`), Choices (`choices`), and Answer (`answer`): These parts should
remain unchanged unless the keys of the tables and the choices and answers for multiple-choice questions.

Here are two examples:
<Two In-context Examples>

Please rewrite the following problem based on the aforementioned requirements and examples:
<Template-based Problem>
    \end{lstlisting}
    \end{tcolorbox}
    \caption{Prompt for paraphrasing template-based problems to natural problems.}
    \label{fig:paraphrase}
\end{figure}

Though ensuring correctness, the template-based TMWPs lack contextual backgrounds, which limits their applicability in practical scenarios. To address this issue, we use an LLM to paraphrase these problems into a more natural and contextual form, improving their generality without sacrificing the original data and solution logic. As shown in Figure \ref{fig:paraphrase}, the prompts for the paraphrase process include instructions, three guidelines for each component of a problem (question, table, answer, and solution), two in-context examples, and a template-based problem. After the paraphrase process, we filter out problems whose answers obtained by the solution are inconsistent with those calculated by the template. We also remove questions in the test set where at least one sample has a BLEU score greater than $\delta$ to prevent potential data leakage issues.

In summary, our approach ensures both the correctness and diversity of problems compared with previous work. First, we classify and abstract the problems into a class of templates and randomly generate a series of questions, tables, answers, and solutions through a rigorous algorithmic procedure. By regarding the template-based problems as effective supervision, this approach avoids hallucination when generating problems with LLMs. At the same time, we introduce different contexts to the problems through the LLM with high language understanding and generation capabilities, enhance the complexity and authenticity of the problems, and ensure that the description of the problems and solutions conforms to the expression habits of English. In general, our method combines the accuracy of program algorithms in generating mathematical problems with the flexibility of LLMs so that high-quality and diverse TMWPs can be generated on a large scale. In addition, our method can be easily extended to new and unseen types of questions, showing its robustness and adaptability.

\section{Experiments}
\label{sec:exp}

\subsection{Baselines}

We compare our performance against the following baselines: (1) \textit{Heuristic Baselines} include heuristic guess and human performance. (2) \textit{Fine-tuned LMs}: We consider UnifiedQA, TAPEX, and TaCo under the fine-tuning setting to predict the final answers. \textbf{UnifiedQA} \cite{khashabi-etal-2020-unifiedqa} is a T5-based model pre-trained on $8$ question answering datasets of multiple formats. \textbf{TAPEX} \cite{liu2022tapex} is a BART-based TaLM pre-trained on tabular data. \textbf{TaCo} \cite{zheng-etal-2023-chain} coordinates two separate TaLMs for CoT generation and answer inference, respectively. We select the large version for the models. (3) \textit{Few-shot Prompting} includes \textbf{GPT-3} \cite{NEURIPS2020_1457c0d6} and \textbf{ChatGPT}. (4) \textit{Few-shot CoT Prompting}: We consider standard \textbf{GPT-3}, \textbf{ChatGPT}, and \textbf{GPT-4} \cite{openai2024gpt4technicalreport}. We also select the following models: \textbf{PromptPG} \cite{lu2023dynamic} selects in-context examples for test samples with a policy gradient method. \textbf{PoT} (Program-of-Thoughts) \cite{chen2023program} exploits Codex to generate the text and Python program for mathematical computations, where GPT-4 is used as the backbone model. \textbf{Chameleon} \cite{NEURIPS2023_871ed095} composes various tools, such as LLM, table verbalizer, and program generator, to accomplish the task.

\subsection{Datasets and Evaluation}

\begin{figure}[t!]
    \centering
    \includegraphics[width=0.8\linewidth]{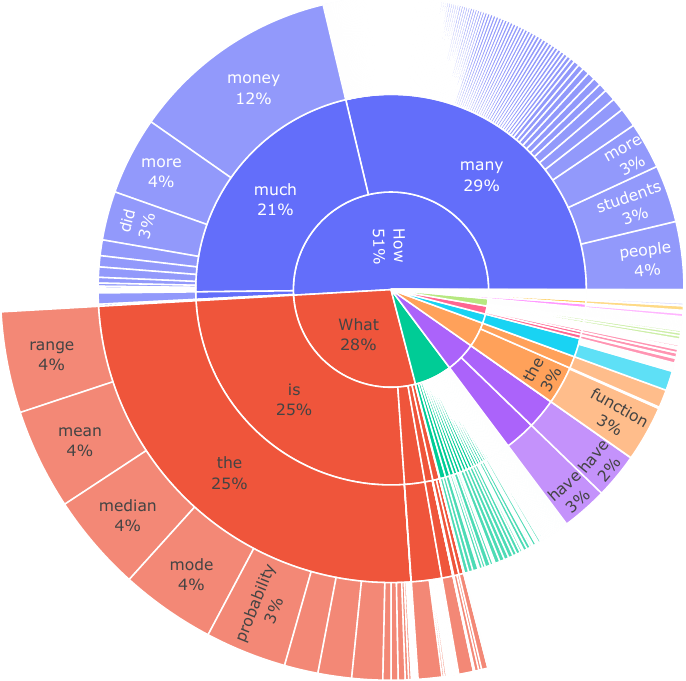}
    \caption{Question distribution of the TabMWP dataset.}
    \label{fig:distribution}
\end{figure}

We conduct evaluations on TabMWP \cite{lu2023dynamic}, a recent large-scale dataset containing $38,431$ grade-level MWPs with tabular context, whose statistics are presented in Table \ref{tab:statistics}. It consists of two types of questions: $28,719$ free-text questions (FREE) with integer (INT) and decimal (DEC) answers, and $9,712$ multiple-choice questions (MC) with extractive text answers (EXTR), Boolean text answers (BOOL) and other text answers (OTH). Apart from the golden answers, the dataset also contains problem solutions in a free-form format. We visualize the distributions of the questions in Figure \ref{fig:distribution}, among which we select $25$ main question types to create our TabMWP-TeLL dataset. For evaluation, we employ exact match accuracy to evaluate overall performance and the performance with respect to each question type, and we adopt the official evaluation script to evaluate the model performance on the test set.

\begin{table}[t!]
    \centering
    \small
    \begin{tabular}{l|cccc}
        \toprule
         & \textbf{Train} & \textbf{Valid} & \textbf{Test} & \textbf{Total} \\
         \midrule
        \#Question & $23,059$ & $7,686$ & $7,686$ & $38,431$ \\
        \#Free-text & $17,135$ & $5,710$ & $5,694$ & $28,719$ \\
        \#MCQ & $5,744$ & $1,976$ & $1,992$ & $9,712$ \\
        \midrule
        \#Table & $22,620$ & $7,546$ & $7,549$ & $37,644$ \\
        \#Solution & $21,623$ & $7,365$ & $7,378$ & $35,442$ \\
        \bottomrule
    \end{tabular}
    \caption{Statistics of the TabMWP dataset.}
    \label{tab:statistics}
\end{table}

\begin{table*}[t!]
    \centering
    \small
    \begin{tabular}{l|cccccccccccccc}
        \toprule
        \multirow{2}{*}{\textbf{Model}} & \multirow{2}{*}{\textbf{\#Para.}} & \multicolumn{2}{c}{\textbf{Question Type}} & & \multicolumn{5}{c}{\textbf{Answer Type}} & & \multicolumn{2}{c}{\textbf{Grade}} & & \multirow{2}{*}{\textbf{Avg.}} \\
        \cmidrule{3-4} \cmidrule{6-10} \cmidrule{12-13}
         & & FREE & MC & & INT & DEC & EXTR & BOOL & OTH & & 1-6 & 7-8 & \\
        \midrule
        \rowcolor{gray!10} & \multicolumn{14}{c}{\textit{Heuristic Baselines}} \\
        \midrule
        Heuristic Guess & $-$ & $6.71$ & $39.81$ & & $8.37$ & $0.26$ & $30.80$ & $51.22$ & $26.67$ & & $17.55$ & $12.27$ & & $15.29$ \\
        Human Perform. & $-$ & $84.61$ & $93.32$ & & $84.95$ & $83.29$ & $97.18$ & $88.69$ & $96.20$ & & $94.27$ & $81.28$ & & $90.22$ \\
        \midrule
        \rowcolor{gray!10} & \multicolumn{14}{c}{\textit{Fine-tuned LMs}} \\
        \midrule
        UnifiedQA$_\textsc{large}$ & $770$M & $48.67$ & $82.18$ & & $55.97$ & $20.26$ & $94.63$ & $68.89$ & $79.05$ & & $65.92$ & $45.92$ & & $57.35$ \\
        TAPEX$_\textsc{large}$ & $400$M & $51.00$ & $80.02$ & & $59.92$ & $16.31$ & $95.34$ & $64.00$ & $73.33$ & & $67.11$ & $47.07$ & & $58.52$ \\
        TaCo$_\textsc{large}$ & $800$M & $91.69$ & $93.47$ & & $92.54$ & $88.41$ & $96.05$ & $91.44$ & $86.67$ & & $92.37$ & $91.86$ & & $92.15$ \\ 
        \midrule
        \rowcolor{gray!10} & \multicolumn{14}{c}{\textit{Few-shot Prompting}} \\
        \midrule
        GPT-3 & $175$B & $54.69$ & $64.11$ & & $58.36$ & $40.40$ & $75.95$ & $52.41$ & $53.02$ & & $63.10$ & $49.16$ & & $57.13$ \\
        ChatGPT & UNK & $65.84$ & $64.61$ & & $66.55$ & $63.09$ & $74.67$ & $54.67$ & $55.24$ & & $69.75$ & $59.88$ & & $65.52$ \\
        \midrule
        \rowcolor{gray!10} & \multicolumn{14}{c}{\textit{Few-shot CoT Prompting}} \\
        \midrule
        Mistral & $7$B & $47.32$ & $48.34$ & & $47.62$ & $46.17$ & $48.30$ & $48.87$ & $44.09$ & & $47.66$ & $47.49$ & & $47.59$ \\
        Qwen 2 & $7$B &  $67.84$ & $68.44$ & & $68.06$ & $66.94$ & $68.12$ & $69.49$ & $62.56$ & & $67.94$ & $68.06$ & & $67.99$ \\
        Llama 3 & $8$B & $71.87$ & $72.34$ & & $72.35$ & $70.00$ & $72.36$ & $72.91$ & $67.12$ & & $72.36$ & $71.51$ & & $71.99$ \\
        GPT-3 & $175$B & $60.76$ & $69.09$ & & $60.04$ & $63.58$ & $76.49$ & $69.19$ & $67.30$ & & $68.62$ & $55.31$ & & $62.92$ \\
        ChatGPT & UNK & $80.89$ & $87.50$ & & $79.36$ & $86.87$ & $81.86$ & $94.00$ & $84.76$ & & $82.68$ & $82.51$ & & $82.60$ \\
        GPT-4 & $1.8$T$^*$ & $90.81$ & $88.48$ & & $97.49$ & $86.16$ & $97.51$ & $96.86$ & $99.11$ & & $89.52$ & $92.40$ & & $88.70$ \\
        PromptPG & $175$B & $66.17$ & $74.11$ & & $64.12$ & $74.16$ & $76.19$ & $72.81$ & $65.81$ & & $71.20$ & $64.27$ & & $68.23$ \\
        PoT (GPT-4) & $1.8$T$^*$ & $97.40$ & $95.58$ & & $98.48$ & $93.22$ & $96.25$ & $98.00$ & $68.57$ & & $96.97$ & $96.87$ & & $96.93$ \\
        Chameleon & $1.8$T$^*$ & $98.95$ & $98.29$ & & $99.34$ & $97.42$ & $98.58$ & $98.56$ & $93.33$ & & $98.95$ & $98.54$ & & $98.78$ \\
        \midrule
        \rowcolor{gray!10} & \multicolumn{14}{c}{\textit{Fine-tuned LLMs (Trained with TabMWP)}} \\
        \midrule
        Mistral & $7$B & $90.87$ & $97.79$ & & $90.93$ & $90.64$ & $97.77$ & $99.44$ & $83.81$ & & $92.67$ & $92.65$ & & $92.66$ \\
        Qwen 2 & $7$B & $92.10$ & $97.84$ & & $92.54$ & $90.39$ & $96.76$ & $99.67$ & $92.38$ & & $94.24$ & $92.71$ & & $93.59$ \\
        Llama 3 & $8$B & $94.13$ & $94.73$ & & $94.81$ & $91.50$ & $98.78$ & $90.11$ & $96.19$ & & $96.61$ & $91.19$ & & $94.29$ \\
        \midrule
        \rowcolor{gray!10} & \multicolumn{14}{c}{\textit{Ours (Trained with TabMWP + TabMWP-TeLL)}} \\
        \midrule
        Mistral & $7$B & $96.08$ & $98.14$ & & $96.69$ & $93.73$ & $98.07$ & $99.22$ & $89.52$ & & $96.77$ & $96.42$ & & $96.62$ \\
        Qwen 2 & $7$B & $97.07$ & $97.94$ & & $97.39$ & $\mathbf{95.79}$ & $97.47$ & $\mathbf{99.44}$ & $89.52$ & & $97.22$ & $97.39$ & & $97.29$ \\
        Llama 3 & $8$B & $\mathbf{97.91}$ & $\mathbf{98.54}$ & & $\mathbf{98.56}$ & $95.36$ & $\mathbf{98.58}$ & $99.33$ & $\mathbf{91.43}$ & & $\mathbf{98.57}$ & $\mathbf{97.42}$ & & $\mathbf{98.07}$ \\
        \bottomrule
    \end{tabular}
    \caption{Experimental results of our models trained with TabMWP and TabMWP-TeLL against baselines. ``Human Perform.'' denotes human performance, ``UNK'' denotes unknown, and * denotes the estimated size of GPT-4 \cite{bambhaniya2024demystifying}.}
    \label{tab:results}
\end{table*}

\subsection{Experimental Setup}

We perform template augmentation and problem paraphrasing with Yi (\texttt{Yi-Large-Turbo}), an LLM that has close performance GPT-4 but with high cost-effectiveness. We access the model via its official API\footnote{\url{https://platform.01.ai/}}, and we set the threshold $\delta$ for the BLEU score as $0.95$. In main experiments, we fine-tune three commonly used LLMs, including Mistral-7B \cite{jiang2023mistral7b}, Qwen 2-7B \cite{yang2024qwen2technicalreport}, and Llama 3-8B \cite{dubey2024llama3herdmodels} with the training set of TabMWP and TabMWP-TeLL, and the evaluations are conducted on the TabMWP test set. During the fine-tuning process, we set the number of epochs as $2$, the batch size per device as $12$, the gradient accumulation steps as $4$, and the learning rate as $2e-4$. To achieve parameter-efficient fine-tuning, we adopt the QLoRA strategy \cite{NEURIPS2023_1feb8787} with XTuner\footnote{\url{https://github.com/InternLM/xtuner}}. All experiments are conducted on $8$ NVIDIA GeForce RTX 3090 graphics cards.

\subsection{Main Results}

Table \ref{tab:results} illustrates the comparison of our models jointly trained with TabMWP and TabMWP-TeLL against baselines. By analyzing the experimental results, we have the following observations: 

(1) The fine-tuned LLMs outperform almost all baselines, except Chameleon, built with the most advanced, closed-source GPT-4. The performance hierarchy among different approaches is clearly established, with few-shot prompting being the least effective, followed by few-shot CoT prompting, and culminating in the fine-tuned LLMs demonstrating the highest efficacy. This gradient of performance highlights the significant impact of large-scale, high-quality training data and reasoning steps on model capability in solving TMWPs.

(2) After incorporating TabMWP-TeLL, the performance of LLMs could be further enhanced. Specifically, Mistral, Qwen 2, and Llama 3 obtain an increase of $3.96\%$, $3.70\%$, and $3.78\%$, respectively. Among the experimented models, Llama 3 demonstrates substantial improvements and outperforms human performance by $7.85\%$, close to the performance of Chameleon by a significantly narrow margin. However, it only has $0.4\%$ of parameters and without deliberated tools compared with Chameleon, underscoring the critical role of strategic augmentation of training data in elevating LLMs' reasoning performance.

(3) TabMWP-TeLL plays a critical role in improving performance on challenging problems while maintaining performance on simple ones.
Initially, LLMs demonstrate strong capabilities in handling simple questions but struggle significantly with more challenging ones. Taking GPT-3 as an example, it achieves an accuracy of $68.62\%$ on problems in grades 1-6 but only $55.31\%$ on questions in problems 7-8. However, by jointly training with TabMWP-TeLL, LLMs, like Llama 3, mark substantial improvements in answering difficult questions (i.e., $97.42\%$) without compromising on simpler ones (i.e., $98.57\%$). This finding emphasizes the importance of high-quality data augmentation in enhancing the robustness and versatility of LLMs, thereby ensuring more consistent performance across a diverse range of question complexities.

\subsection{Ablation Study}

\begin{table}[t!]
    \centering
    \small
    \begin{tabular}{l|cccc}
        \toprule
        \textbf{Model} & \textbf{FREE} & \textbf{MC} & \textbf{Overall} & $\mathbf{\Delta}$ \\
        \midrule
        Mistral & $90.87$ & $97.79$ & $92.66$ & $-$ \\
        \quad \textit{w/ Template} & $91.41$ & $98.29$ & $93.20$ & $0.54$ \\
        \quad \textit{w/ Ques.+Table} & $91.27$ & $97.64$ & $92.92$ & $0.26$ \\
        \quad \textit{w/ Ques. Only} & $92.48$ & $\mathbf{98.64}$ & $94.08$ & $1.42$ \\
        \quad \textit{Ours} & $\mathbf{96.08}$ & $98.14$ & $\mathbf{96.62}$ & $\mathbf{3.96}$ \\        
        \midrule
        Qwen 2 & $92.10$ & $97.84$ & $93.59$ & $-$ \\
        \quad \textit{w/ Template} & $92.20$ & $\mathbf{98.19}$ & $93.75$ & $0.16$ \\
        \quad \textit{w/ Ques.+Table} & $92.48$ & $97.79$ & $93.86$ & $0.27$ \\
        \quad \textit{w/ Ques. Only} & $93.47$ & $97.84$ & $94.60$ & $1.01$ \\
        \quad \textit{Ours} & $\mathbf{97.07}$ & $97.94$ & $\mathbf{97.29}$ & $\mathbf{3.70}$ \\
        \midrule
        Llama 3 & $94.13$ & $94.73$ & $94.29$ & $-$ \\
        \quad \textit{w/ Template} & $93.08$ & $97.99$ & $94.35$ & $0.06$ \\
        \quad \textit{w/ Ques.+Table} & $93.26$ & $98.24$ & $94.55$ & $0.26$ \\
        \quad \textit{w/ Ques. Only} & $94.33$ & $98.69$ & $95.46$ & $1.17$ \\
        \quad \textit{Ours} & $\mathbf{97.91}$ & $\mathbf{98.54}$ & $\mathbf{98.07}$ & $\mathbf{3.78}$ \\
        \bottomrule
    \end{tabular}
    \caption{Comparison of different TMWP generation methods. $\mathbf{\Delta}$ calculates the overall improvement.}
    \label{tab:ablation}
\end{table}

\paragraph{Effectiveness of TeLL for Problem Generation.} We first analyze the effectiveness of TeLL, the template-driven, LLM-paraphrased TMWP generation strategy by comparing it with three additional methods: generating template-based problems, utilizing LLMs to generate new questions and tables, and utilizing LLMs to only rewrite the questions. To ensure a fair comparison, all three methods generate $23$K problems. Experimental results are organized in Table \ref{tab:ablation}. The template-based problems perform the worst due to their inability to adapt to real questions with diverse backgrounds. The generated questions and tables with LLMs show subpar performance due to inaccuracies that occur within. Besides, when using LLMs only to rewrite questions, despite some performance improvements, they are limited by the unchanged numbers within the questions and tables. In contrast, our proposed method, which generates questions based on templates and LLMs, effectively balances correctness and diversity in the augmented questions, leading to superior results on TMWP solving.

% \begin{figure}[t!]
%     \centering
%     \includegraphics[width=\linewidth]{AnonymousSubmission/Figures/StepByStep.pdf}
%     \caption{Comparison between different training data to the performance on the TabMWP dataset.}
%     \label{fig:step-by-step}
% \end{figure}

\paragraph{Effectiveness of Illustrative Solutions.}

\begin{table}[t!]
    \centering
    \small
    \begin{tabular}{l|ccc}
        \toprule
        \textbf{Solution} & \textbf{Mistral} & \textbf{Qwen 2} & \textbf{Llama 3} \\
        \midrule
        Free-form & $92.66$ & $93.59$ & $94.29$ \\
        Illustrative & $\mathbf{94.29}$ & $\mathbf{95.86}$ & $\mathbf{96.21}$ \\
        \bottomrule
    \end{tabular}
    \caption{Performance of models trained on free-form and illustrative solutions.}
    \label{tab:illustrative}
\end{table}

We then analyze the efficacy of the illustrative solutions compared with the original free-form solutions for TMWP solving by fine-tuning the LLMs with the same questions but with different solution types, i.e., free-form solutions and the illustrative solutions, in which the illustrative solutions are constructed by LLMs given the free-form solutions. As shown in Table 
\ref{tab:illustrative}, the model trained with free-form solutions performs worse due to the lack of detail and structure. Our proposed illustrative solutions ultimately produce the best performance, indicating that the step-by-step solutions offer a more systematic and detailed approach, thus significantly improving the model performance.

\paragraph{Effectiveness across Multiple Question Types.}

\begin{table}[t!]
    \centering
    \small
    \begin{tabular}{c|cccc}
        \toprule
        \textbf{Type} & \textbf{\#Problem} & \textbf{TableMWP} & \textbf{TeLL} & \textbf{Combination} \\
        \midrule
        3 & $273$ & $66.67$ & $83.88$ & $94.14$ \\
        6 & $501$ & $69.46$ & $73.65$ & $96.51$ \\
        8 & $502$ & $68.92$ & $83.27$ & $95.82$ \\
        10 & $100$ & $100.00$ & $100.00$ & $100.00$ \\
        11 & $110$ & $100.00$ & $100.00$ & $100.00$ \\
        22 & $266$ & $98.12$ & $96.24$ & $98.12$ \\
        23 & $280$ & $100.00$ & $99.64$ & $100.00$ \\
        25 & $281$ & $100.00$ & $99.64$ & $100.00$ \\
        \midrule
        Avg. & $2313$ & $82.49$ & $88.24$ & $97.45$ \\
        \bottomrule
    \end{tabular}
    \caption{Experimental results of Llama 3 trained with TabMWP, TabMWP-TeLL, and the combination of both. }% of both on the TabMWP dataset.}
    \label{tab:types}
\end{table}

We also examine the accuracy of various question types by training Llama 3 on three distinct datasets: TabMWP, TabMWP-TeLL, and a mix of both. As depicted in Table \ref{tab:types}, the models trained solely on TabMWP demonstrate competence in solving straightforward problems, such as calculating mean and median; however, they struggle with more complex tasks, particularly those involving stem-and-leaf plots. For instance, the performance on three types of stem-leaf plots is only $66.67\%$, $69.46\%$, and $68.92\%$, respectively. In contrast, the model trained on TabMWP-TeLL exhibits superior performance, showing an average improvement of $11.92\%$ on the sampled stem-leaf plot problems, attributed to the inclusion of more detailed and illustrative solutions in our generated dataset, which enhanced the models' multi-step reasoning capabilities.

\paragraph{Effects of the Size of Augmented Data.}

\begin{figure}[t!]
    \centering
    \includegraphics[width=0.9\linewidth]{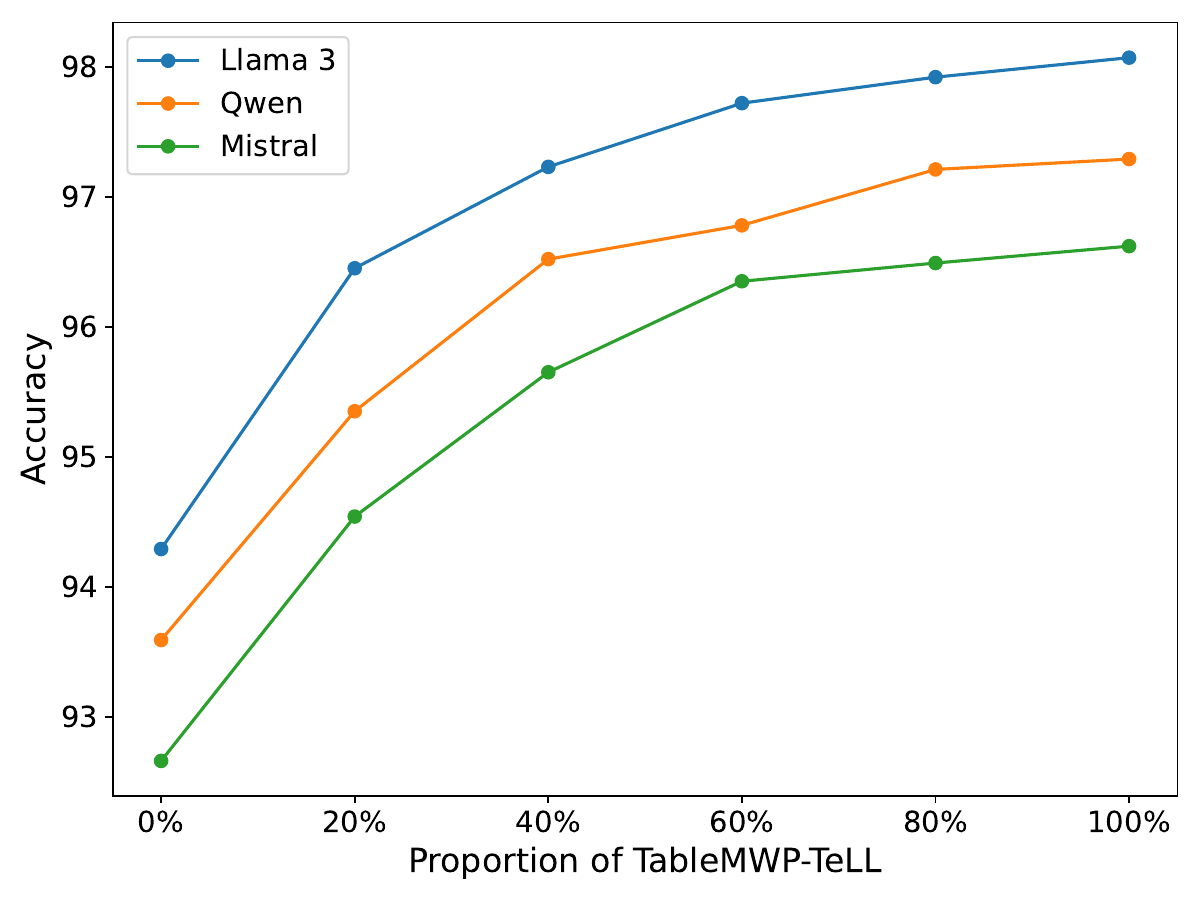}
    \caption{Effects of the proportion of TabMWP-TeLL.}% on TMWP-solving performance.}
    \label{fig:samples}
\end{figure}

We further investigate the effects of the size of augmented TabMWP-TeLL data with the TabMWP dataset on LLMs' TMWP-solving performance. As shown in Figure \ref{fig:samples}, incorporating more TabMWP-TeLL data consistently contributes to the performance across all models, even with smaller amounts of augmented data. Among these, a small amount of data (e.g., $20\%$) leads to a substantial improvement in model performance. As the data volume increases, the improvement rate slows, but the model continues to benefit from the generated data. This finding not only demonstrates the usefulness of additional augmented data in enhancing TMWP results but also highlights the critical importance of high-quality data in achieving superior model performance.

\paragraph{Human Verification.}

To guarantee the quality of our generated TMWPs, we sample $1,000$ examples from TabMWP-TeLL for human verification. Generally, the generated data achieve high correctness, with a rate of $97.5\%$. We categorize the generation errors into three primary types during the problem paraphrasing step: incomplete paraphrased questions (LLMs may generate questions that are incomplete and unanswerable), incorrect paraphrased solutions (LLMs may generate solutions that are not logically correct with hallucination issues), and grammatical errors. We will address these errors in our future work.

\section{Conclusion and Future Work}
\label{sec:conclusion}

We introduce TeLL, a template-driven, LLM-paraphrased framework to generate high-quality TMWPs by leveraging both templates and LLMs. The framework consists of five steps, including abstraction, augmentation, selection, instantiation of templates, and the paraphrasing of problems. We construct a high-quality dataset called TabMWP-TeLL by adhering to the question types in the TabMWP dataset with illustrative step-by-step solutions, and we conduct experiments with three commonly used LLMs. Experimental results demonstrate the effectiveness of TabMWP-TeLL in improving TMWP-solving by incorporating it with TabMWP, making LLMs' performance outperform the baselines, and it is particularly effective in improving performance on challenging problems while maintaining the performance on simple ones. In the future, we will generalize our method to more complex reasoning tasks, such as commonsense reasoning and multi-hop reasoning.

\section{Acknowledgments}
We thank all anonymous reviewers for their valuable comments. This work was partially supported by the following: National Natural Science Foundation of China under No. 92370119, 62436009 and 62376113, XJTLU Funding REF-22-01-002, RDF-TP-0019, and Suzhou Municipal Key Laboratory for Intelligent Virtual Engineering (SZS2022004).

\bibliography{aaai25}

\onecolumn
\section{Question Types in TabMWP-TeLL}
\label{sec:intro}

Table \ref{tab:ques_types} outlines the diverse question types included in the TabMWP-TeLL dataset. It encompasses $25$ distinct types of TMWPs, which have been curated from the TabMWP dataset. Specifically, problem types 1-11 correspond to stem-leaf plot problems, types 12-17 are related to trading scenarios, types 18-19 involve multiple-choice questions, types 20-21 focus on probability, and types 22-25 cover arithmetic operations. These categories represent a broad spectrum of mainstream TMWPs, ensuring substantial diversity within the dataset. Additionally, the proposed template-driven, LLM-paraphrased framework can also be easily generalized to unseen question types.

\begin{table*}[!htbp]
    \centering
    \begin{tabular}{m{1cm}<{\centering}|m{14cm}}
        \toprule
        \textbf{Type} & \textbf{Template} \\
        \midrule
        1 & How many times does \verb|{count_value}| appear in the stem-and-leaf plot? \\
        2 & How many numbers are at least \verb|{range_start}| and at most \verb|{range_end}|? \\
        3 & How many numbers are at least \verb|{range_start}| but fewer than \verb|{range_end}|? \\
        4 & How many numbers are greater than \verb|{range_start}| but fewer than \verb|{range_end}|? \\
        5 & How many numbers are greater than \verb|{range_start}| and at most \verb|{range_end}|? \\
        6 & How many numbers are fewer than \verb|{threshold}|? \\
        7 & How many numbers are at most \verb|{threshold}|? \\
        8 & How many numbers are at least \verb|{threshold}|? \\
        9 & How many numbers are greater than \verb|{threshold}|? \\
        10 & What is the smallest number in the dataset? \\
        11 & What is the largest number in the dataset? \\
        12 & How much money does \verb|{name}| need to buy \verb|{number}| \verb|{products}|? \\
        13 & How much money does \verb|{name}| need to buy \verb|{number1}| \verb|{product1s}| and \verb|{number2}| \verb|{product2s}|? \\
        14 & How much money does \verb|{name}| need to buy \verb|{number1}| \verb|{product1s}|, \verb|{number2}| \verb|{product2s}|, and \verb|{number3}| \verb|{product3s}|? \\
        15 & How much money will \verb|{name}| have left if \verb|{gender}| buys \verb|{number}| \verb|{products}|? \\
        16 & How much money will \verb|{name}| have left if \verb|{gender}| buys \verb|{number1}| \verb|{product1s}| and \verb|{number2}| \verb|{product2s}|? \\
        17 & How much money does \verb|{name}| have left if \verb|{gender}| buys \verb|{number1}| \verb|{product1s}|, \verb|{number2}| \verb|{product2s}|, and \verb|{number3}| \verb|{product3s}|? \\
        18 & Which category has more value for \verb|{column}|, \verb|{row1}| or \verb|{row2}|? \\
        19 & Which category has less value for \verb|{column}|, \verb|{row1}| or \verb|{row2}|? \\
        20 & What is the probability that a randomly selected item is \verb|{row}| and \verb|{col}|? \\
        21 & What fraction of \verb|{items}| in the table belong to \verb|{category}|? \\
        22 & What is the mean of the numbers? \\
        23 & What is the median of the numbers? \\
        24 & What is the mode of the numbers? \\
        25 & What is the average of the numbers? \\
        \bottomrule
    \end{tabular}
    \caption{Question types covered in the TabMWP-TeLL dataset.}
    \label{tab:ques_types}
\end{table*}

\end{document}